\documentclass[letterpaper, 10 pt, conference]{icra-sub/ieeeconf}  

\IEEEoverridecommandlockouts                              

\usepackage{amsmath}
\usepackage{mathtools}
\usepackage{amssymb}
\usepackage{amsfonts}   
\usepackage{dsfont}

\usepackage[pdftex]{graphicx}
\usepackage{wrapfig}
\usepackage{caption}
\usepackage{placeins}

\usepackage{booktabs}   
\usepackage{makecell}

\usepackage{algorithm}
\usepackage{algorithmic}

\usepackage[table]{xcolor}  
\usepackage{hyperref}
\hypersetup{
    colorlinks=true,
}

\usepackage[]{mdframed}

\usepackage{cleveref}

\usepackage{xparse}     
\usepackage{expl3}      
\usepackage{xspace}
\usepackage{lipsum}     

\crefname{section}{Sec.}{Secs.}
\crefname{figure}{Fig.}{Figs.}
\crefname{table}{Tab.}{Tabs.}
\crefname{equation}{Eq.}{Eqs.}

\DeclarePairedDelimiterX{\infdivx}[2]{(}{)}{%
  #1\;\delimsize\|\;#2%
}

\ExplSyntaxOn
\newcommand\latinabbrev[1]{
  \peek_meaning:NTF . {%
    \textit{#1}\@}%
  { \peek_catcode:NTF a {%
      \textit{#1}.\@ }%
    {\textit{#1}.\@}}}
\ExplSyntaxOff

\newif\ifARXIVversion
\ARXIVversiontrue

\def\MethodAcronym{KinoGen\xspace}

\title{\LARGE \bf \MethodAcronym: Customizable Humanoid Loco-Manipulation References via Kinematically Grounded Video Generation}

\ifARXIVversion
\author{%
Qi Wu$^{1}$, Lulu Zhao$^{1}$, Sanghyun Hahn$^{1}$,
Rishi Yennu$^{1}$, Emmy Yang$^{1}$, Dingqi Zhang$^{2}$, Tao Pang,\\
Jiuguang Wang$^{2}$, Kuan Fang$^{1}$\\[0.5em]
{\small $^{1}$Cornell University \qquad $^{2}$Multiply Labs}%
}
\else
\author{Anonymous Authors}
\fi

\begin{document}
\bstctlcite{BSTcontrol}

\ifARXIVversion
\IEEEaftertitletext{\vskip-1em\relax}
\fi
\maketitle
\thispagestyle{empty}
\pagestyle{empty}
\bibliographystyle{icra-sub/bib/IEEEtran}

\begin{abstract}
We present \MethodAcronym{}, a framework that synthesizes customizable humanoid loco-manipulation references through kinematically grounded video generation. \MethodAcronym{} conditions a pretrained video generator on calibrated renders of the target robot and object in a simulated task instance and recovers the reference by fitting the same models to the generated two-view video, so that the reference lies in the robot configuration space by construction without a human body proxy or retargeting. Body, hand, and object priors initialize the reconstruction, and direct optimization against cross-view evidence refines it. Since an edited keyframe is simply another rendered state, a user can prescribe a goal object pose, regenerate the video around the edited keyframe, and obtain a reference that satisfies the pose exactly. Across eight loco-manipulation tasks, \MethodAcronym{} yields lower penetration and foot skating than video-based baselines and preserves over 90\% of annotated contacts, and its customized references satisfy every prescribed target with substantially less penetration than direct interpolation. Tracking policies trained on the synthesized references are deployed on a Unitree G1 humanoid with dexterous hands.
Project webpage: \url{https://kinogen-humanoid.github.io/}.

\end{abstract}
\section{Introduction}
\label{sec:intro}

Humanoid robots can learn impressive whole-body skills by tracking a reference motion with reinforcement learning. 
Whereas a locomotion reference specifies only the body motion~\cite{luo2025sonic, liao2025beyondmimic}, a loco-manipulation reference must specify synchronized whole-body, hand, and object trajectories. Such references must vary with the interaction behavior and the scene layout, and each must match the kinematics of the target robot and the geometry of its task. Many tasks further require the object to reach a prescribed pose. Obtaining such references remains a central bottleneck. Human demonstrations must be reconstructed and retargeted to the robot, which introduces embodiment errors, and teleoperation requires substantial human effort to cover the variation in behaviors, objects, scenes, and goals.

Video generators offer a scalable alternative: text and image conditioning can produce diverse, scene-conditioned interaction behaviors without manually collecting every demonstration \cite{jang2025dreamgen,li2025novaflow}. 
For humanoids, recent methods either generate human--object interactions and retarget the recovered motion to the robot~\cite{ni2025genmimic,xie2026grail} or generate the target robot directly~\cite{bi2026genhoi,xu2026dream2act}. For the first family, since the behavior is generated and reconstructed through a human body proxy, the robot's kinematics and the exact task instance constraints can only be satisfied after generation, which can cause penetration and drifting. The second family still relies on human-motion recovery and retargeting or a separately trained robot-specific pose detector, and recovers neither hands nor objects. Consequently, these pipelines do not directly recover complete whole-body, hand, and object references through the robot and object models that define the task instance.
Moreover, text- or image-conditioned generation is open loop with respect to metric task requirements. 
The outcome may depict the intended behavior but cannot be revised to satisfy a prescribed configuration.
We need references that are generated for the target robot in the exact task instance and can be customized to satisfy a prescribed configuration.

\begin{figure}[t]
    \centering
    \vspace{7pt}
    \includegraphics[width=\linewidth]{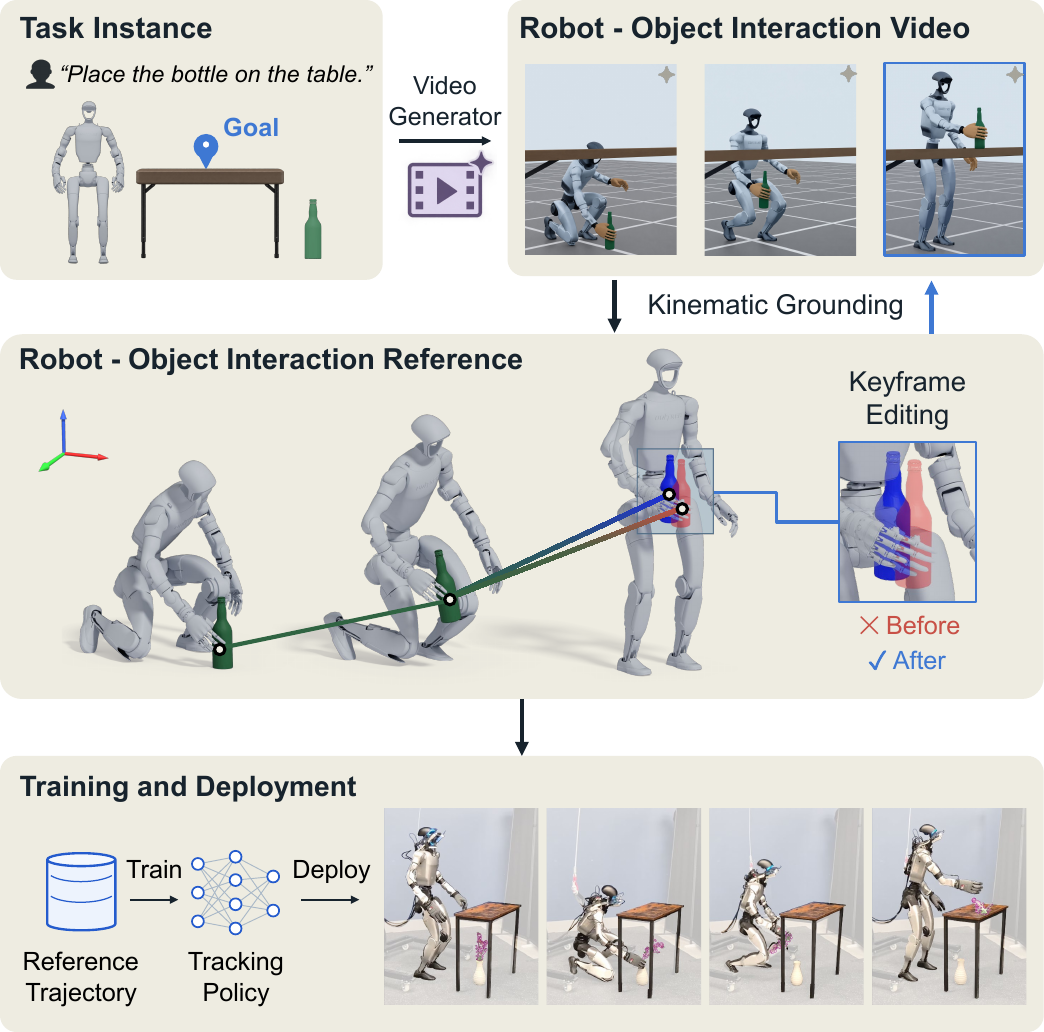}
    \caption{
      \textbf{KinoGen} synthesizes and customizes humanoid loco-manipulation references for the target embodiment.
    A task specification and calibrated simulator render condition robot-video
    generation. Fitting the target robot and object models to the generated video
    recovers synchronized whole-body, hand, and object trajectories. Editing a
    robot--object keyframe conditions regeneration of the intervening motion, while
    constrained reconstruction enforces the correction. The resulting references
    train tracking policies deployed on a Unitree G1 with dexterous hands.
    }
    \vspace{-10pt}
    \label{fig:teaser}
\end{figure}

Our key insight, as shown in Fig.~\ref{fig:teaser}, is that a pretrained video generator and a physics simulator can be coupled through rendering and reconstruction using the same robot and object models. The generator supplies the interaction behavior, and the simulator supplies the exact embodiment and task geometry. Rendering the simulated task instance into the conditioning image places the generated behavior in the target embodiment and scene, and fitting the same models to the generated video returns the behavior to the simulator. We refer to this coupling as kinematic grounding. It has two consequences. The complete body, hand, and object reference is recovered directly in the robot configuration space and the object pose space by construction, without a human body proxy, retargeting, or a pose detector. In addition, a keyframe edited in simulation is simply another rendered state, so the same interface lets the generator regenerate the behavior around the edit while reconstruction enforces it exactly.
 
In this work, we present \textbf{\MethodAcronym{}}, a framework that synthesizes customizable humanoid loco-manipulation references through kinematically grounded video generation. \MethodAcronym{} first grounds generation by conditioning the video generator on renders of the task instance from two calibrated views, so that the generated clip depicts the target robot with dexterous hands in the specified scene. It then reconciles the generated video with body, hand, and object priors to initialize reconstruction with plausible motion. Next, it fits the robot and object models to both views by direct optimization in the configuration space to recover the complete reference. Finally, a user may prescribe a goal object pose, and \MethodAcronym{} edits the corresponding keyframe in simulation, renders it back into the generator, and enforces it in constrained reconstruction to obtain a reference that satisfies the pose exactly.

We evaluate \MethodAcronym{} on eight loco-manipulation tasks for a Unitree G1 humanoid with dexterous hands, spanning sitting, kneeling, mopping, and the lifting of boxes, baskets, chairs, dumbbells, and basketballs. Against three video-based reference pipelines, \MethodAcronym{} yields the lowest penetration and foot skating, preserves over 90\% of annotated contacts, and trains tracking policies with the highest success rate. Ablations show that the second view, point tracks, and motion priors each contribute to accurate and coherent reconstruction. In reference customization, \MethodAcronym{} satisfies every prescribed target while incurring substantially less penetration than direct interpolation. Finally, tracking policies trained on the synthesized references execute four of the tasks on the physical robot.

\section{Related Work}

\noindent\textbf{Humanoid loco-manipulation.}
Humanoid whole-body control has advanced rapidly through reinforcement learning and imitation from human motion \cite{luo2025sonic,yin2025unitracker}. 
Meanwhile, dexterous humanoid loco-manipulation couples high-dimensional whole-body motion with contact-rich object interaction, requiring coordinated robot--object references \cite{zhang2024wococo, yang2025omniretarget, weng2025hdmi}. Previous methods use references commonly originated from teleoperation or human motion capture \cite{fu2025humanplus,he2024omnih2o, ze2025twist2} which limits coverage to captured interactions and can introduce retargeting errors. 

\noindent\textbf{Video generators for robot data synthesis.}
Video generators have been used as world models and a scalable source of demonstrations from which motion references can be recovered~\cite{jang2025dreamgen,li2025novaflow}. 
For object-centric manipulation, RIGVid~\cite{patel2025rigvid} extracts a 6D object trajectory from a generated demonstration.
For humanoid data synthesis, recent approaches either generate human or proxy interactions and retarget the recovered motion~\cite{ni2025genmimic,zhou2026exoactor,xie2026grail}, or depict the target robot but recover its motion through human-motion retargeting, a robot-specific pose detector, or sparse point guidance ~\cite{bi2026genhoi,xu2026dream2act,chen2026imagine2real}. GRAIL~\cite{xie2026grail} grounds its proxy in the scene, but still requires retargeting to recover robot motion.
Large-scale robot-video models such as Uni-Mo~\cite{liu2026unimo} and
OmniHumanoid~\cite{song2026omnihumanoid} further demonstrate that embodiment-conditioned generation can scale across robots, but do not recover complete robot--object reference trajectories.
In contrast, \MethodAcronym{} conditions generation on calibrated renders of the simulator state of the task instance, and fits those same models to multi-view evidence, following a render-and-compare formulation~\cite{labbe2021robopose}. This directly recovers coordinated body, hand, and object trajectories, without a human body proxy or a separately trained robot-specific pose detector.

\noindent\textbf{Customizable video generation and editing.}
Generated videos may express the intended task without satisfying a prescribed metric configuration. Prior work controls generated motion using event keyframes, edge maps, rendered trajectories, or cross-embodiment editing \cite{he2026skip,chen2026craft,kim2026robotfactored,guo2026roboedit}; other methods treat the video as fixed and ground its output through downstream optimization or simulation \cite{han2026video2sim2real,ziakas2026grounding,he2026vdreamer}. 
GRAIL mentions terminal-frame conditioning but does not evaluate intermediate edits. 
None of these methods regenerates motion from an edited reconstructed robot--object state and then enforces that state exactly in the recovered reference. Through kinematic grounding, \MethodAcronym{} closes this loop by rendering an edited reconstructed robot--object keyframe for keyframe-conditioned regeneration and enforcing the complete keyframe as a hard constraint during reconstruction.

\begin{figure*}[t]
    \centering
    \includegraphics[width=\linewidth]{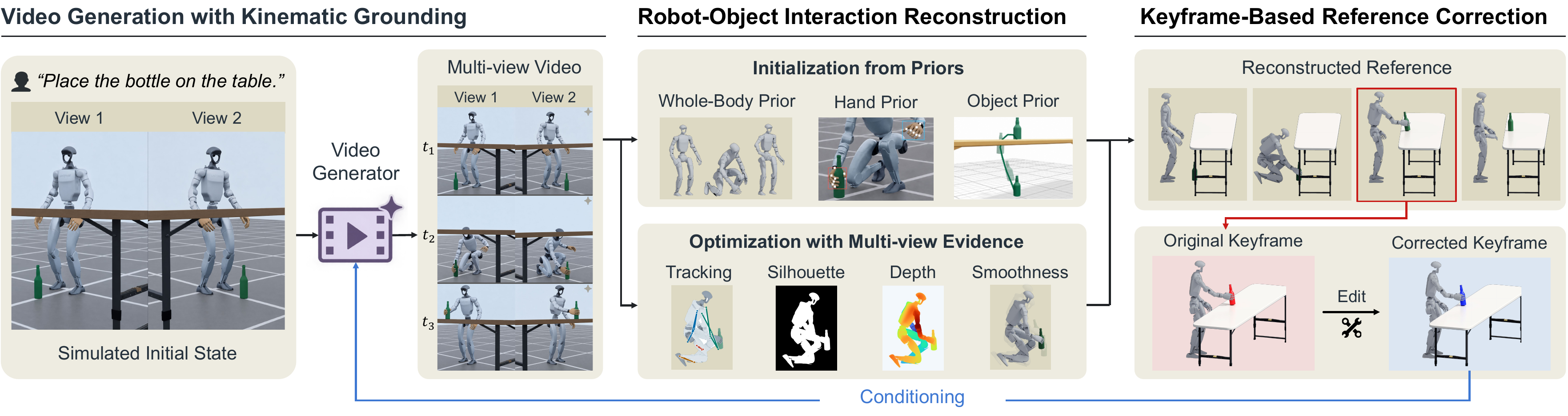}
    \caption{\textbf{Overview of \MethodAcronym{}.}  
    \emph{Left:} A task prompt and calibrated renders of the initial robot--object
    state condition synchronized multi-view video generation
    (Sec.~\ref{sec:robot_video_generation}).
    \emph{Middle:} Motion priors are reconciled with the
    generated observations to initialize reconstruction
    (Sec.~\ref{sec:motion_prior}); direct optimization then fits the robot and object configurations to multi-view evidence
    (Sec.~\ref{sec:robot-object reconstruction}).
    \emph{Right:} Reference correction enforces prescribed robot–object configurations through keyframe editing, keyframe-conditioned regeneration, and constrained reconstruction (Sec.~\ref{sec:metric_constraints}).}
    \vspace{-10pt}
    \label{fig:method-data}
\end{figure*}
\section{Method}
\label{sec:method}

We present \MethodAcronym{}, a framework that synthesizes customizable humanoid loco-manipulation references through kinematically grounded video generation, as summarized in Fig.~\ref{fig:method-data}. Central to our approach is the bidirectional interplay between a pretrained video generator and a physics simulator through rendering and reconstruction. The generator supplies the interaction behavior, and the simulator supplies the exact embodiment and task geometry. \MethodAcronym{} thereby combines the generative prior of the video model with the kinematic constraints of the simulator to synthesize diverse and customizable references for learning loco-manipulation skills.

Formally, a task instance comprises the target robot and object models, the scene geometry, the initial robot--object configuration $q_0$, and a task prompt $\tau$. The goal is to synthesize a robot--object trajectory $q_{1:T}$ over $T$ frames in the simulator world frame that serves as the loco-manipulation reference, where $q_t=(q_t^{\mathrm r},q_t^{\mathrm o})$, $q_t^{\mathrm r}\in\mathcal{Q}^{\mathrm r}$ comprises the floating-base pose, body joint angles, and finger coordinates, and $q_t^{\mathrm o}\in\mathrm{SE}(3)$ is the object pose. To correct a reference, a user may additionally prescribe a goal object pose $q^{\mathrm o}_{k,\mathrm{goal}}$ at frame $k$ that the reference must satisfy exactly. Applying the pipeline to $N$ task instances yields a reference dataset $\mathcal{D}=\{q^{(i)}_{0:T_i}\}_{i=1}^{N}$. The references in $\mathcal{D}$ serve as targets for a reference-tracking policy $\pi$, which learns to reproduce them under simulated physics before deployment.

\MethodAcronym{} proceeds in four stages. We first ground generation by conditioning the video generator on renders of the target robot itself (Sec.~\ref{sec:robot_video_generation}). Then, we reconcile the generated video with a robot motion prior to initialize reconstruction with plausible motion (Sec.~\ref{sec:motion_prior}). Next, we return the behavior to the simulator by fitting the robot and object models to both views, without a human proxy or a pose detector (Sec.~\ref{sec:robot-object reconstruction}). Finally, we close the loop by rendering an edited keyframe back into the generator and enforcing it in reconstruction (Sec.~\ref{sec:metric_constraints}).

\subsection{Video Generation with Kinematic Grounding}
\label{sec:robot_video_generation}

\MethodAcronym{} begins by grounding video generation in the simulator, as shown in Fig.~\ref{fig:method-data} (left). From a task instance, including the initial configuration $q_0$ and the task prompt $\tau$, we seek a video with frames $\hat{I}_t$ for $t=1,\dots,T$ in which the target robot performs the prompted task in the specified scene. A pretrained generator does not produce such a video from the prompt alone, since the prompt leaves the embodiment, the scene, and the initial configuration unspecified. Even once these are specified, a single view leaves motion along the viewing ray ambiguous, especially for the hands and the object. The rendered robot hands, moreover, differ from the natural imagery that pretrained video and hand models expect. We resolve each of these issues through how the initial state is rendered into the conditioning image, as described next.

\noindent\textbf{Conditioning on the rendered initial state.} We condition generation on an image of the initial state rendered in the simulator, $\mathcal{R}_c(q_0)$, where $\mathcal{R}_c(q)$ renders the known robot and object models in configuration $q$ from calibrated camera $c$. The render preserves the target robot geometry and configuration, the object geometry and pose, the scene layout, and the camera viewpoint. Conditioning on it places the generated behavior in the same embodiment and spatial setting as the desired reference and removes the need to align a free-form clip with the simulator afterward. The rendered image thus provides a shared metric reference between generation and reconstruction, and the same operator $\mathcal{R}_c$ renders every candidate configuration during reconstruction (Sec.~\ref{sec:robot-object reconstruction}).

\noindent\textbf{Cross-view consistency.} Since the conditioning image is rendered in simulation, adding a view requires no additional data collection. We therefore render the initial state from two calibrated cameras, $\mathcal{R}_1(q_0)$ and $\mathcal{R}_2(q_0)$, and concatenate the two $640\times720$ views into a $1280\times720$ conditioning image. Generating both views within one clip encourages a temporally aligned interaction across viewpoints. Each generated frame $\hat{I}_t$ thus contains both views, which we split into $\hat{I}_{t,c}$ for $c\in\{1,2\}$ as the synchronized multi-view evidence for reconstruction.
In practice, when the generated views exhibit a large cross-view inconsistency, such as depicting different moving limbs, we reject the clip and rerun generation with a different random seed.

\noindent\textbf{Appearance editing.}
The hands occupy the smallest image regions and are later processed by a prior trained on human hands. We therefore recolor only the hand surfaces with a consistent skin-toned material, which alters their appearance but not their geometry.
\ifARXIVversion
Meanwhile, we leave the remaining robot surfaces unchanged.
\fi
The render $\mathcal{R}_c(q_0)$ remains the ground truth for the masks and robot surface points used in reconstruction.

With the conditioning image in place, we use Seedance~2.0~\cite{seedance2026} to generate a short clip from it and a prompt that specifies the task, static cameras, and cross-view consistency. Each clip lasts five to ten seconds at 24~fps. The resulting synchronized two-view clip is passed to Sec.~\ref{sec:motion_prior}, which extracts calibrated visual evidence and initializes reconstruction.

\subsection{Reconciling Video and Motion Priors}
\label{sec:motion_prior}

The generated video specifies the desired interaction visually but does not provide the 3D state that a reference requires. We therefore convert it into calibrated observations and initializations for reconstruction, as shown in Fig.~\ref{fig:method-data} (middle). The two-view frames $\hat{I}_{t,c}$ are not associated with the robot model, and their depth carries no metric scale. We thus begin by extracting instance masks, robot surface-point tracks, and metric depth that tie every frame and view to the known geometry. These observations alone do not determine the motion. Framewise fitting of the redundant humanoid can match the images while producing incoherent or implausible motion, robot masks only weakly constrain the hands, and frequent hand--object occlusion makes direct object fitting sensitive to initialization. We therefore reconcile body, hand, and object priors with both views to initialize the body, hand, and object trajectories.

\noindent\textbf{Calibrated visual evidence.}
From each generated frame $\hat{I}_{t,c}$, we extract instance masks $M^{e}_{t,c}$ for the entities $e\in\{\mathrm r,\mathrm o,\mathrm L,\mathrm R\}$, robot surface-point tracks $\mathbf{u}_{t,c,j}$ indexed by surface point $j$, and metric depth $D_{t,c}$, which together link the generated views to the known robot, object, and camera geometry. After splitting the two views, we fold all crop and resize transformations into the calibrated camera intrinsics. We initialize the generic segmentation model SAM~3~\cite{carion2025sam3} with the exact frame-$0$ robot, object, and hand masks and propagate each instance independently. From the known frame-$0$ robot configuration, we project 650 visible robot surface points per view and propagate their image positions using a generic point tracker TAPNext++~\cite{jung2026tapnextpp}. 
This associates image measurements directly with the target robot model without training a robot-specific keypoint or pose detector. 
Points are initialized only at frame~$0$; tracks that become occluded or inconsistent with the robot mask are marked invalid. 
Because monocular depth has unknown metric scale, we estimate temporally consistent relative depth with Video Depth Anything~\cite{chen2025videodepthanything} and calibrate it using analytic intersections between camera rays and the ground plane. The robot and object pixels are excluded from the fit. The resulting instance masks, robot surface-point tracks, and metric depth maps are used to reconcile the priors below and for cross-view reconstruction in Sec.~\ref{sec:robot-object reconstruction}.

\noindent\textbf{Whole-body motion prior.}
We initialize the body trajectory with Kimodo~\cite{rempe2026kimodo}, a text-conditioned motion model 
trained to produce G1 trajectories directly in the robot configuration space. The same task description conditions both the video generator and this model.
Let $\widetilde q^{\mathrm r}(\epsilon)=G_{\theta}(\epsilon;\tau,q_0^{\mathrm r})$ be its frozen denoising process, conditioned on the task prompt $\tau$ and the known initial robot configuration.  Following Diffusion Noise Optimization (DNO)~\cite{karunratanakul2024dno}, we backpropagate the multi-view observation loss through the denoising trajectory and update only its initial noise.
\ifARXIVversion
We run DNO for 200 iterations, sampling 12 frames per iteration.
\fi
This objective reconciles the body prior with the multi-view evidence while keeping the optimized noise close to a valid sample:
\begin{equation}
\begin{aligned}
    \epsilon^\star=\arg\min_\epsilon\quad
    &\mathcal{L}_{\mathrm{obs}}\!\left(\widetilde q^{\mathrm r}(\epsilon)\right)
    +\lambda_{\mathrm{stat}}\mathcal{L}_{\mathrm{stat}}(\epsilon)\\
    &+\lambda_{\mathrm{trust}}\|\epsilon-\epsilon_0\|_2^2 .
    \label{eq:dno_body}
\end{aligned}
\end{equation}
Here $\mathcal{L}_{\mathrm{obs}}$ evaluates the silhouette, depth, and ground terms of Eq.~\eqref{eq:body_objective} on sampled frames. The noise-statistics loss $\mathcal{L}_{\mathrm{stat}}(\epsilon) = \mu(\epsilon)^2 + (\sigma(\epsilon)-1)^2$, where $\mu(\epsilon)$ and $\sigma(\epsilon)$ are the mean and standard deviation of the entries of $\epsilon$, drives these moments toward those of a standard normal, while the trust-region term keeps it close to the initial sample $\epsilon_0 \sim \mathcal{N}(0,I)$.
Thus, noise optimization selects a task-consistent trajectory as the body prior for whole-body reconstruction.

\noindent\textbf{Hand and object pose priors.}
We initialize the wrist orientation and finger articulation of each hand with the image-based hand estimator WiLoR~\cite{potamias2024wilor} and triangulate its palm position from the two views. This hand estimate is the only quantity in the pipeline obtained with a human body model, and it serves only as initialization.
\ifARXIVversion
Each 11-joint robot hand is represented by three coupled coordinates: thumb flexion, shared flexion of the remaining fingers, and thumb abduction-. The hand prior initializes them with rule-based mappings. Direct fitting subsequently refines all three. 
\fi
For the object, we run the pose tracker FoundationPose~\cite{wen2024foundationpose} independently in both views and select the candidate with the best cross-view agreement. These observations and initializations are the inputs to cross-view reconstruction in Sec.~\ref{sec:robot-object reconstruction}.

\subsection{Cross-View Reconstruction of Robot-Object Interaction}
\label{sec:robot-object reconstruction}

Reconstruction returns the generated behavior to the simulator by fitting the robot and object models to both views, as shown in Fig.~\ref{fig:method-data} (middle). From the observations and initializations of Sec.~\ref{sec:motion_prior}, it recovers the complete reference $q_{1:T}$. Because we optimize directly in the robot configuration space and the object pose space, the recovered reference lies in these spaces by construction. Every image term renders the current estimate with $\mathcal{R}_c$ and compares it with the masks, tracks, and depth extracted from $\hat{I}_{t,c}$, following a render-and-compare formulation~\cite{labbe2021robopose}. We fit the hands, the body, and the object in sequence rather than jointly, because the views observe each of them differently. The small image region of a hand cannot reliably drive the complete robot chain, so each hand is first solved as a floating base. The whole body is then solved from the reconciled body prior, with the finger coordinates fixed and the wrists constrained to the hand solution. The object, which the hands frequently occlude, is solved last with the reconstructed robot fixed, so that object-pose ambiguity cannot corrupt the robot motion and occlusion by the solved parts is accounted for. Each solve optimizes its configuration variables directly with Adam.

\noindent\textbf{Floating-base hand reconstruction.}
Treating each hand as a floating base decouples wrist-pose and finger-configuration estimation from the complete robot chain. For hand $s\in\{\mathrm L,\mathrm R\}$, let $\eta_{t,s}=(H_{t,s},f_{t,s})$ denote its floating-base state, comprising the wrist pose $H_{t,s}\in\mathrm{SE}(3)$ and the finger coordinates $f_{t,s}$, and let $\mathcal{R}_c(\eta_{t,s})$ render hand $s$ alone. Starting from the hand prior in Sec.~\ref{sec:motion_prior}, we optimize $\eta_{1:T,s}$ so that multi-view masks align the robot hand with the observed hand regions, while the hand-prior and temporal terms stabilize frames with weak or occluded evidence:

\begin{equation}
\begin{aligned}
    \mathcal{L}_{\mathrm{hand}}(\eta_{1:T,s})={}&\sum_{t,c} w_{t,c,s}\mathcal{L}_{\mathrm{mask}}\!\left(\mathcal{R}_c(\eta_{t,s}),M^{s}_{t,c}\right)\\
    &+\lambda_m\mathcal{L}_{\mathrm{hand\text{-}prior}}+\lambda_a\mathcal{L}_{\mathrm{acc}} .
    \label{eq:floating_hand}
\end{aligned}
\end{equation}
\ifARXIVversion
Here, $w_{t,c,s}$ weights each view by its valid hand-mask area normalized by its 85th-percentile area over time, with invalid masks assigned zero weight. 
\fi
$\mathcal{L}_{\mathrm{mask}}$ combines target-normalized false-negative and false-positive areas with IoU after object and body depth clip the rendered hand; $\mathcal{L}_{\mathrm{hand\text{-}prior}}$ penalizes deviation from the hand-prior initialization, and $\mathcal{L}_{\mathrm{acc}}$ penalizes temporal acceleration.
The optimized wrist poses $H_{t,s}$ and finger coordinates $f_{t,s}$ then serve as targets for whole-body reconstruction.

\noindent\textbf{Whole-body reconstruction with hand targets.}
Recovering the complete robot trajectory $q^{\mathrm r}_{1:T}$ draws on all three kinds of evidence. Silhouettes constrain projected extent, tracks disambiguate internal articulation, and depth constrains motion along the viewing rays. We combine these cues with regularization:
\begin{equation}
    \begin{aligned}
      \mathcal{L}_{\mathrm{body}}(q_{1:T}^{\mathrm r})={}&
      \lambda_{\mathrm{sil}}\mathcal{L}_{\mathrm{chamfer}}
      +\lambda_{\mathrm{trk}}\mathcal{L}_{\mathrm{track}}
      +\lambda_d\mathcal{L}_{\mathrm{depth}} \\
      &+\lambda_g\mathcal{L}_{\mathrm{ground}} 
      +\lambda_{\mathrm{lim}}\mathcal{L}_{\mathrm{limit}}
      +\lambda_{\mathrm{temp}}\mathcal{L}_{\mathrm{temp}} \\
      &+\lambda_{\mathrm{prior}}\mathcal{L}_{\mathrm{prior}} .
      \label{eq:body_objective}
    \end{aligned}
\end{equation}
Here, $\mathcal{L}_{\mathrm{chamfer}}$ is the Chamfer distance between the contours of $\mathcal{R}_c(q_t^{\mathrm r})$ and of the robot mask $M^{\mathrm r}_{t,c}$. $\mathcal{L}_{\mathrm{track}}$ penalizes the reprojection error of each robot surface point $j$ against its track $\mathbf{u}_{t,c,j}$, and $\mathcal{L}_{\mathrm{depth}}$ compares the rendered depth with $D_{t,c}$. The remaining terms discourage ground penetration, joint-limit violations, and temporal jitter, and $\mathcal{L}_{\mathrm{prior}}$ penalizes deviation from the body-prior trajectory $\widetilde q^{\mathrm r}(\epsilon^\star)$.  

To preserve the recovered hand motion during whole-body reconstruction, we
initialize from $\widetilde q^{\mathrm r}(\epsilon^\star)$, fix the finger
coordinates to $f_{t,s}$, and optimize
\begin{equation}
    \mathcal{L}_{\mathrm{robot}}(q_{1:T}^{\mathrm r})
    =
    \mathcal{L}_{\mathrm{body}}(q_{1:T}^{\mathrm r})
    +
    \lambda_{\mathrm{wrist}}
    \mathcal{L}_{\mathrm{wrist}}(q_{1:T}^{\mathrm r};H,\omega).
    \label{eq:robot_hand_refinement}
\end{equation}
The wrist loss robustly penalizes the position and orientation differences between the robot wrists and the hand-pose targets $H=\{H_{t,s}\}$, with each target weighted by its multi-view observation reliability $\omega_{t,s}=\sum_c w_{t,c,s}$ collected in $\omega$. The optimized trajectory is fixed
during the subsequent object reconstruction.

\noindent\textbf{Rigid-object reconstruction.}
Recovering the object trajectory completes the reference. With the reconstructed robot and hands fixed, we initialize from the object prior in Sec.~\ref{sec:motion_prior} and optimize $q^{\mathrm o}_{1:T}$, where multi-view silhouette and depth terms align the object with the visual evidence and velocity and acceleration terms regularize its trajectory:
\begin{equation}
\mathcal{L}_{\mathrm{obj}}(q_{1:T}^{\mathrm o})
=
\lambda_o\mathcal{L}^{\mathrm{obj}}_{\mathrm{chamfer}}
+\lambda_d\mathcal{L}^{\mathrm{obj}}_{\mathrm{depth}}
+\lambda_v\mathcal{L}^{\mathrm{obj}}_{\mathrm{vel}}
+\lambda_a\mathcal{L}^{\mathrm{obj}}_{\mathrm{acc}} .
\label{eq:object_refinement}
\end{equation}
The initial object pose $q_0^{\mathrm o}$ remains fixed at its known value. Here, $\mathcal{L}^{\mathrm{obj}}_{\mathrm{chamfer}}$ compares the projected object samples with the object mask $M^{\mathrm o}_{t,c}$, and $\mathcal{L}^{\mathrm{obj}}_{\mathrm{depth}}$ compares their depths with $D_{t,c}$. The weights $\lambda$ are set separately for each objective.
The resulting $q_{1:T}^{\mathrm o}$, together with the fixed robot trajectory $q_{1:T}^{\mathrm r}$, forms the complete robot--object reference $q_{1:T}$. 
We perform no subsequent joint refinement of the robot, hands, and object.

\subsection{Reference Correction via Keyframe Editing}
\label{sec:metric_constraints}

Reference customization closes the loop between the simulator and the generator, as shown in Fig.~\ref{fig:method-data} (right) and Fig.~\ref{fig:metric_constraint_comparison}. Given a reconstructed reference $q_{1:T}$ and a goal object pose $q^{\mathrm o}_{k,\mathrm{goal}}$ prescribed at frame $k$, we produce a customized reference that satisfies the prescribed pose exactly. The goal pose alone leaves the whole-body configuration at frame $k$ unspecified, so we first complete it into a robot--object keyframe $q_k^{\mathrm{goal}}=(q_{k,\mathrm{goal}}^{\mathrm r},q_{k,\mathrm{goal}}^{\mathrm o})$ in simulation. A keyframe in turn constrains only one instant, and interpolating states toward it may cross obstacles or support surfaces. We therefore render the keyframe back into the generator, which serves as a \emph{behavior-aware interpolator} that synthesizes the intervening task-conditioned motion. Since video generation alone does not guarantee metric accuracy, reconstruction finally enforces $q_k=q_k^{\mathrm{goal}}$ exactly. Conditioning generation on keyframes has been explored for controlling generated robot motion~\cite{he2026skip,xie2026grail}. What distinguishes our use is that the keyframe is an edited robot--object state rendered from the simulator and enforced in the recovered reference.

\noindent\textbf{Keyframe editing.}
Given the prescribed object pose $q_{k,\mathrm{goal}}^{\mathrm o}$, we derive the robot configuration $q_{k,\mathrm{goal}}^{\mathrm r}$ of the keyframe as follows. Starting from the reconstructed configuration $q_k^{\mathrm r}$, we transfer the object-pose change to the grasping-hand wrist pose, preserving the hand--object relative transform and finger coordinates. A weighted single-frame inverse-kinematics objective enforces this wrist-pose target while assigning larger preservation weights to the free hand and both feet, producing the complete edited keyframe $q_{k}^{\mathrm{goal}}$ for regeneration.

\noindent\textbf{Keyframe-conditioned regeneration.}
We render the edited state through both calibrated cameras as $\mathcal{R}_c(q_k^{\mathrm{goal}})$ using the same skin-toned hand material as in the initial generation. To preserve the original task phase and timing in our experiments, regeneration is conditioned on the original initial frame, the edited frame, and the last frame of the original clip, together with the task prompt $\tau$. We use Ray~3.2~\cite{lumalabs2026ray32}, which supports multi-keyframe conditioning, to generate the intervening behavior. The regenerated clip is then passed to reconstruction, with its masks, depth maps, and surface-point tracks recomputed.

\noindent\textbf{Constrained reconstruction.}
Because video generation alone does not guarantee metric accuracy, we reconstruct the regenerated clip using the same objectives as before while fixing $q_k=q_k^{\mathrm{goal}}$. The robot temporal loss in Eq.~\eqref{eq:body_objective} and the object velocity and acceleration regularization in Eq.~\eqref{eq:object_refinement} smooth the neighboring configurations around frame $k$. Repeating editing, regeneration, and constrained reconstruction with different edits produces multiple customized references from one grounded clip. The hard kinematic constraint guarantees the prescribed configuration, but does not separately guarantee contact or physical executability.

\begin{figure*}[t]
    \centering
    \includegraphics[width=\linewidth]{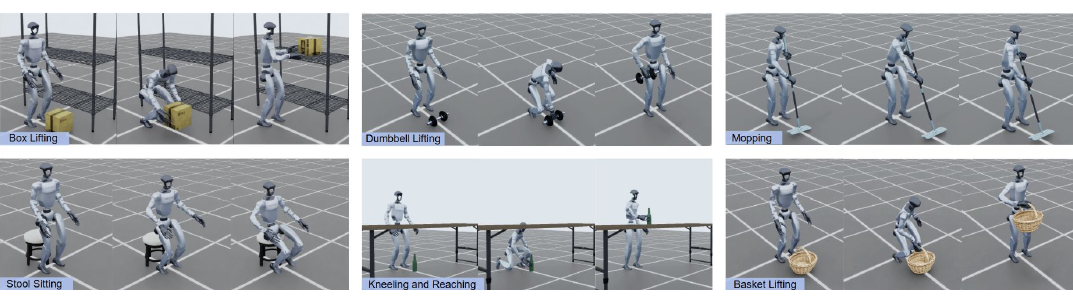}
    \caption{\textbf{Synthesized robot--object references.}  We show box lifting, dumbbell lifting,
    floor mopping, stool sitting, kneeling and reaching, and basket lifting, all produced by the same pipeline.
    Each panel contains temporally ordered keyframes from one generated and
    reconstructed trajectory.}
    \label{fig:DiverseData}
\end{figure*}

\begin{table*}[h]
    \centering
    \caption{Quantitative comparison of generated humanoid reference quality and downstream RL performances}
    \label{tab:baseline_comparison}
    \setlength{\tabcolsep}{6pt}
    \begin{tabular*}{\linewidth}
    {@{\extracolsep{\fill}}lccccccc@{}}
        \toprule
        & \multicolumn{4}{c}{Reconstruction quality}
        & \multicolumn{3}{c}{Downstream RL Policy} \\
        \cmidrule(lr){2-5}\cmidrule(lr){6-8}
        Method
        & \makecell{Penetration\\{\scriptsize (cm)} $\downarrow$}
        & \makecell{Foot skating\\{\scriptsize (cm/s)} $\downarrow$}
        & \makecell{Contact preservation\\{\scriptsize (\%)} $\uparrow$}
        & \makecell{Smoothness\\{\scriptsize (m/s$^2$)} $\downarrow$}
        & \makecell{Success\\{\scriptsize (\%)} $\uparrow$}
        & \makecell{Reference deviation\\{\scriptsize (m)} $\downarrow$}
        & \makecell{Learning curve AUC\\{\scriptsize } $\uparrow$} \\
        \midrule
        GenMimic\cite{ni2025genmimic} & 3.17 & 9.95 & 13.57 & 0.96 & 31 & 0.16 & 29.92 \\
        GenHOI\cite{bi2026genhoi}   & 10.55 & 6.52 & 43.91 & 1.34 & 11 & 0.15 & 19.90 \\
        GRAIL\cite{xie2026grail}    & 3.06 & 5.11 & 69.40 & \textbf{0.59} & 42 & 0.13 & 38.73 \\
        \midrule
        \MethodAcronym{} (ours) & \textbf{2.30} & \textbf{4.09} & \textbf{91.16} & 0.73 & \textbf{57} & \textbf{0.09} & \textbf{60.74} \\
        \bottomrule
    \end{tabular*}
    \vspace{-10pt}

\end{table*}

\section{Experiments}

Our experiments address three questions. \textbf{Q1:} does kinematically grounded video generation produce higher-quality references than generated-human reconstruction? \textbf{Q2:} which components are necessary for accurate and coherent robot--object reconstruction? \textbf{Q3:} can keyframe customization satisfy a prescribed configuration exactly while avoiding the penetrations caused by direct interpolation?

\subsection{Experiment Setup}
\noindent\textbf{Task suite and simulation.}
We test our method on a Unitree G1 humanoid equipped with a pair of 6-DoF OYMotion ROHand dexterous hands. The scene is rendered in Isaac Sim. To evaluate whole-body interactions with different postures, contacts, object geometries, and manipulation goals, our task suite contains eight tasks: sitting on a stool; kneeling to reach an object; floor mopping; lifting a box from the ground and placing it on a shelf; and lifting a basket, chair, dumbbell, and basketball. Together, these tasks cover large posture changes and contact with the environment. For each task, we instantiate the scene in simulation, generate the interaction video, and reconstruct trajectories. Figure~\ref{fig:DiverseData} shows the resulted robot--object references across task instances and interaction modes.

\noindent\textbf{Policy training and evaluation.}
To evaluate the downstream utility of each reconstructed reference $q_{0:T}$, we train a state-based reference-tracking policy in simulation. Following DeepMimic-style tracking~\cite{peng2018deepmimic}, the policy conditions on robot proprioception, object states and the reference phase, and outputs target joint commands for the low-level controller. 
Its reward tracks the reference robot and object motion and the hand--object relative transform, with regularization for smooth control. The policy architecture, observations, reward, simulator settings, training procedure, and training budget are identical for every reference source with only the reference trajectories differing. 
We report task success if the robot doesn't fall and the position tracking error is less than 0.2\,m. We also report reference deviation measured as robot bodies tracking error, and the area under the reward-versus-environment-step curve divided by total iteration. 
We train with 8,192 parallel environments for 6,000 PPO updates and repeat 10 times with different seeds to obtain the averaged results.

\begin{figure*}[t]
    \centering

    \begin{minipage}[t]{0.625\textwidth}
        \vspace{0pt}
        \centering
        \includegraphics[
        width=\linewidth,
            trim={0pt 2pt 0pt 0pt},
        clip
        ]
            {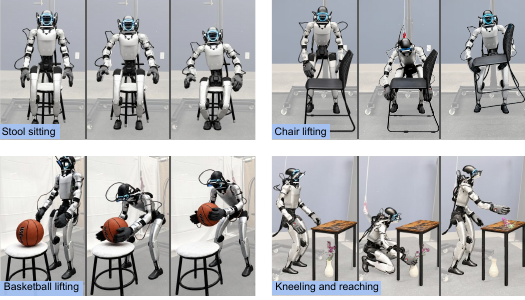}
        \caption{\textbf{Hardware deployment of reconstructed references.}
        Tracking policies execute four \MethodAcronym{} references on a Unitree G1:
        stool sitting, chair lifting, basketball lifting, and kneeling and reaching.
        Frames within each panel progress from left to right and show one hardware
        rollout.}
        \label{fig:real_world_deployment}
    \end{minipage}
    \hfill
    \begin{minipage}[t]{0.35\textwidth}
        \vspace{0pt}
        \centering
        \includegraphics[
            width=\linewidth,
            trim={55pt 0pt 55pt 0pt},
            clip
        ]{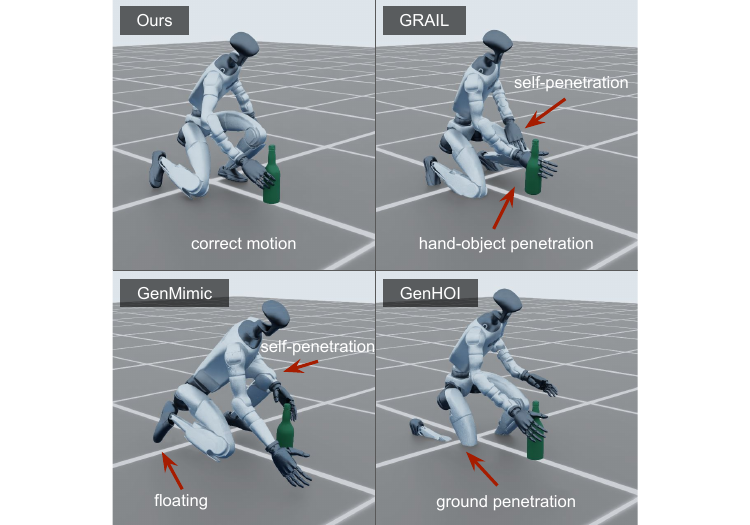}
        \caption{\textbf{Qualitative reconstruction comparison.}
        \MethodAcronym{} preserves grasp and ground contact, whereas baselines exhibit penetration, floating, or misalignment.}
        \label{fig:baseline_comparison}
    \end{minipage}

    \vspace{-10pt}
\end{figure*}

\noindent\textbf{Hardware deployment.}
As shown in Fig.~\ref{fig:real_world_deployment}, we qualitatively deploy the reference-tracking policies on four hardware tasks: stool sitting, chair pickup, basketball pickup, and object pickup from a kneeling posture. The policies are trained with privileged full-state observations and distilled for deployment into policies conditioned on robot proprioception and the initial object pose. The object pose is estimated once from the first RGB-D frame using FoundationPose~\cite{wen2024foundationpose}. 
\ifARXIVversion
For tasks like stool sitting where the object is invisible from the first frame, the policy reuses the initial object pose from simulation.
\fi
For additional experimental details, results, and videos, please visit our project page: \href{https://kinogen-humanoid.github.io/}{https://kinogen-humanoid.github.io/}.

\subsection{Comparative Results}
To investigate whether kinematically grounded generation and reconstruction produces higher-quality references than human or partial reconstruction, we compare against two types of video-based humanoid reference pipelines. GenMimic~\cite{ni2025genmimic} and GRAIL~\cite{xie2026grail} reconstruct generated human interactions from retargeting. GenHOI~\cite{bi2026genhoi} conditions video generation on a rendered target-robot scene, but uses a human pose estimation model to recover a human motion sequence and then retarget it to the humanoid. In contrast, our method directly reconstructs the complete robot, dexterous-hand, and object trajectories in their target configuration spaces from calibrated multi-view visual evidence. 

\begin{table}[!t]
    \centering
    \caption{\textbf{Ablation of calibrated visual evidence and motion priors.}
    Results are averaged across all evaluation sequences. Each sequence is repeated 10 times.}
    \label{tab:reconstruction_ablation}
    \setlength{\tabcolsep}{4pt}
    \begin{tabular*}{\linewidth}
    {@{\extracolsep{\fill}}lcccc@{}}
        \toprule
        Variant
        & \makecell{Worst IoU\\ $\uparrow$}
        & \makecell{EPE\\(px) $\downarrow$}
        & \makecell{Acceleration\\(m/s$^2$) $\downarrow$}
        & \makecell{Penetration\\(mm) $\downarrow$} \\
        \midrule
        Single view              & 0.42 & 13.80 & 0.47 & 10.4 \\
        w/o point tracks         & \textbf{0.65} & 9.89 & 0.39 & \textbf{10.0} \\
        w/o body prior         & 0.52 & 11.59 & 1.01 & 26.8 \\
        w/o hand prior      & 0.62 & 8.02 & 0.79 & 24.3 \\
        \midrule
        Full model               & \textbf{0.65} & \textbf{7.92} & \textbf{0.37} & 10.5 \\
        \bottomrule
    \end{tabular*}
    \vspace{-10pt}
\end{table}

\ifARXIVversion
To place these baselines on the same task instances, we follow GRAIL's scene-conditioning protocol for GenMimic and GRAIL: in each initial simulator render, we replace the robot with a robot-proportioned human character while keeping the environment fixed.  We then generate one human--object interaction video and provide the same clip to GenMimic and GRAIL. Each method completes reconstruction and retargeting according to its own pipeline.  Our method and GenHOI instead generate the corresponding target-robot video from the same task instance.  Thus, the comparison preserves scene and task content while evaluating the different human-retargeting and direct-robot reconstruction strategies.  
\fi
We report the following metrics: total penetration aggregates robot--object, self-, and robot--ground penetration; foot skating measures the horizontal velocity of planted feet; contact preservation is the fraction of annotated contact frames with a hand--object distance below 2\,cm; and
smoothness measures root translational acceleration.
\ifARXIVversion
For each penetration type, we multiply the mean depth over penetrating frames by the fraction of frames containing that penetration. Contact frames are obtained from VLM-reviewed annotations of the corresponding generated source video. Smoothness is computed as the mean magnitude of the second temporal derivative of the root position.
\fi

\ifARXIVversion
Table~\ref{tab:baseline_comparison} shows that \MethodAcronym{} produces the most physically consistent and task-aligned references overall. It achieves the lowest total penetration and foot-skating velocity, while recovering 91.16\% of annotated contact frames, compared with 69.40\% for the strongest baseline. GRAIL produces the lowest root acceleration, but \MethodAcronym{} remains smoother than GenMimic and GenHOI while providing substantially better contact preservation and physical feasibility, which is aligned with the qualitative illustration in Fig. ~\ref{fig:baseline_comparison}. These improvements also translate to downstream control: policies trained on our references achieve the highest success rate, the lowest reference deviation, and the largest learning-curve AUC. Together, the results indicate that directly reconstructing the target embodiment from kinematically grounded video yields references that are both geometrically coherent and more useful for policy learning. 
\fi

\subsection{Ablation Study}
\ifARXIVversion
We remove one component at a time while keeping all other inputs,
optimization settings, and loss weights fixed.  The \emph{single-view}
variant uses only the right camera; the \emph{w/o point tracks} variant removes
the TAPNext++ reprojection loss; the \emph{w/o body prior} variant replaces the
DNO-optimized motion proposal with initialization from the default pose; and
the \emph{w/o hand prior} variant initializes the wrist and finger trajectories from
the default hand configuration.  We report worst-view mask IoU over the two
calibrated views, mean 2D end-point error (EPE) on surface-point tracks held out from optimization, root translational acceleration for smoothness, and maximum robot--ground penetration during interaction frames.
\fi

Table~\ref{tab:reconstruction_ablation} shows that kinematic grounding requires calibrated multi-view evidence, point tracks and motion priors.  Using only one view reduces worst-view IoU and increases EPE, showing that the second view is important for resolving ambiguity.  Without body prior, root acceleration increases from 0.37 to 1.01\,m/s$^2$ and penetration increases from 10.5 to 26.8\,mm, indicating that image fitting alone does not provide coherent whole-body motion.  Removing hand prior also increases acceleration and penetration, as inaccurate hand targets propagate to the arm and body during refinement.  Point tracks reduce EPE while leaving silhouette overlap nearly unchanged, showing that they constrain internal articulation that silhouettes cannot observe. Together, the multi-view evidence determines where the motion occurs, point tracks provide evidence for internal articulation, while the motion priors keep the motion consistent with the common robot motions. These parts ensure that the method grounds the kinematics of the scene while reconstructing the motion.

\subsection{Reference Correction}

\begin{figure}[!t]
    \centering
    \includegraphics[width=\linewidth]{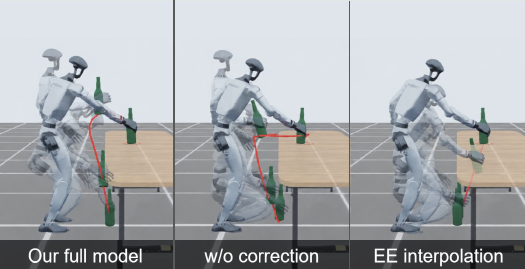}
    \caption{\textbf{Metric reference correction.} 
    \emph{Left:} Keyframe-conditioned regeneration proposes a lift-over motion, and constrained reconstruction enforces the target exactly.
    \emph{Middle:} Reconstructing the original cue-conditioned video does not reach the prescribed target.
    \emph{Right:} Direct end-effector interpolation reaches the target but moves the hand and bottle through the table.
    }
    \label{fig:metric_constraint_comparison}
    \vspace{-5pt}
\end{figure}

\begin{table}[!t]
    \centering
    \caption{\textbf{Metric reference correction.}
    We report the mean target error and the mean per-trial maximum table penetration, taking the worse value between the bottle and operating hand.}
    \label{tab:constraint_ablation}
    \setlength{\tabcolsep}{3pt}
    \begin{tabular*}{\linewidth}
    {@{\extracolsep{\fill}}lcccccc@{}}
        \toprule
        & \multicolumn{3}{c}{Position error (cm) $\downarrow$}
        & \multicolumn{3}{c}{Worst penetration (mm) $\downarrow$} \\
        \cmidrule(lr){2-4}\cmidrule(lr){5-7}
        Variant & Left & Center & Right & Left & Center & Right \\
        \midrule
        w/o correction  & 14.55 & 1.07 & 3.06 & 44.84 & 5.14 & 13.55 \\
        EE interpolation & 0.00 & 0.00 & 0.00 & 139.51 & 147.23 & 139.45 \\
        Full model      & 0.00 & 0.00 & 0.00 & 19.32 & 2.72 & 20.99 \\
        \bottomrule
    \end{tabular*}
    \vspace{-10pt}
\end{table}
We consider a bottle pickup followed by placement on a table.  
A circular decal in the scene marks the desired placement location and provides a visual cue to the video generator.  
We divide the tabletop into left, center, and right regions and sample ten target positions per region.  We use the same original generation for all variants associated with a target. For the customization process, we require the robot to put the bottle precisely at the desired location. 

We compare three variants.  \emph{Without correction} reconstructs the original cue-conditioned video.  \emph{End-effector interpolation} imposes the metric keyframe, preserves the hand--object transform, and directly interpolates the operating-hand translation and orientation between neighboring keyframes before solving per-frame IK.  
\ifARXIVversion
This baseline satisfies the endpoint but has no prior for how to approach it.  
\fi
The \emph{full model} instead regenerates the motion around the metric keyframe before reconstruction and hard constraint enforcement, allowing the video prior to propose a lift-over and approach motion around the table. Figure~\ref{fig:metric_constraint_comparison} illustrates this difference in the intervening motion: direct interpolation crosses the table, whereas keyframe-conditioned generation produces an approach around the support surface. For each region, we report the mean distance between the reconstructed bottle and its target.  
\ifARXIVversion
We quantify collision error as the mean, over trials, of the maximum support-normal translational penetration against the table, taking the worse value between the bottle and operating right hand.  Penetrations within a 2\,mm tolerance are set to zero.  Thus, placement error measures constraint satisfaction, while penetration distinguishes a feasible approach from merely interpolating to the correct endpoint.
\fi

Table~\ref{tab:constraint_ablation} shows that visual conditioning alone does not provide reliable metric control.
\ifARXIVversion
Without correction, the placement error varies from 1.07 to 14.55\,cm across the three regions. 
\fi
Direct end-effector interpolation reaches every target exactly, but produces large penetration because the endpoint does not specify how the robot should approach it. By using this naive interpolation, the robot lifts up the bottle and directly moves through the table. Our full method satisfies every target exactly and reduces the penetration significantly.  The method therefore gives the user precise metric customization over the generated reference, while video generation serves as a behavior interpolator between the grounded keyframes.

\FloatBarrier

\section{Conclusion}
\label{sec:conclusion}
We presented \MethodAcronym{}, a framework for synthesizing customizable humanoid loco-manipulation references through kinematically grounded video generation. By using the same target robot and object models to condition generation and reconstruct motion, \MethodAcronym{} recovers whole-body, dexterous-hand, and object trajectories directly in the target configuration spaces, without full-body human-motion reconstruction or body retargeting. Calibrated multi-view evidence and complementary motion priors reconcile generated behavior with the target embodiment and metric task instance, yielding references with improved physical consistency that support tracking on a Unitree G1. This grounding further enables users to edit robot--object keyframes and use video generation as a behavior-aware interpolator, while constrained reconstruction enforces the prescribed configurations with substantially less penetration than direct interpolation. 

The current design has three limitations. Reconstruction is vulnerable to severe or persistent occlusion, under which masks, point tracks, and depth provide little visual evidence. More broadly, the quality and diversity of the recovered references are bounded by the video generator. Behaviors that are temporally inconsistent, geometrically implausible, or absent from its prior cannot always be recovered through reconstruction. Finally, the pipeline steers the generated behavior only through sparse keyframes, which constrain selected states but leave the intervening motion to the generator. Future work will investigate denser and more interactive feedback between generation and simulation, so that generated videos are continually grounded in the known scene geometry, robot kinematics, and task constraints.




\ifARXIVversion
\section*{ACKNOWLEDGMENT}
We thank Thomas Cohn, Xinghao Zhu, Chong Zhang, Chuanruo Ning, Qianxu Wang, Yunhao Cao, Samuel Jin, Zhen Wu, Zi-ang Cao and Zhongru Zhang for valuable discussions. We thank Xingda Chen for assistance with the hardware setup. We also thank Calvin Qiu and Mengling Zhang for assistance with the experiments.
\fi


\bibliography{reference}

@STRING{tog       = "ACM Transactions on Graphics"}

@IEEEtranBSTCTL{BSTcontrol,
  CTLuse_forced_etal       = "yes",
  CTLmax_names_forced_etal = "1",
  CTLnames_show_etal       = "1"
}

@inproceedings{jang2025dreamgen,
  title={{DreamGen}: Unlocking generalization in robot learning through video world models},
  author={Jang, Joel and Ye, Seonghyeon and Lin, Zongyu and Xiang, Jiannan and Bjorck, Johan and Fang, Yu and Hu, Fengyuan and Huang, Spencer and Kundalia, Kaushil and Lin, Yen-Chen and others},
  booktitle={CoRL},
  year={2025}
}

@article{li2025novaflow,
  title={{NovaFlow}: Zero-shot manipulation via actionable flow from generated videos},
  author={Li, Hongyu and Sun, Lingfeng and Hu, Yafei and Ta, Duy and Barry, Jennifer and Konidaris, George and Fu, Jiahui},
  journal={arXiv:2510.08568},
  year={2025}
}

@article{patel2025rigvid,
  title={Robotic manipulation by imitating generated videos without physical demonstrations},
  author={Patel, Shivansh and Mohan, Shraddhaa and Mai, Hanlin and Jain, Unnat and Lazebnik, Svetlana and Li, Yunzhu},
  journal={arXiv:2507.00990},
  year={2025}
}

@article{liu2026unimo,
  title={Unleashing Infinite Motion: Scaling Expressive Quadrupedal Motion via Generative Video Priors},
  author={Liu, Youzhi and Gao, Li and Qian, Yifei and Liu, Liu and Cai, Yang and Li, Ziqiao},
  journal={arXiv:2606.28237},
  year={2026}
}

@article{song2026omnihumanoid,
  title={{OmniHumanoid}: Streaming Cross-Embodiment Video Generation with Paired-Free Adaptation},
  author={Song, Yiren and Deng, Xiyao and Yang, Pei and Wang, Yihan and Shou, Mike Zheng},
  journal={arXiv:2605.12038},
  year={2026}
}

@article{ni2025genmimic,
  title={From Generated Human Videos to Physically Plausible Robot Trajectories},
  author={Ni, James and Wang, Zekai and Lin, Wei and Bar, Amir and LeCun, Yann and Darrell, Trevor and Malik, Jitendra and Herzig, Roei},
  journal={arXiv:2512.05094},
  year={2025}
}

@article{bi2026genhoi,
  title={{GenHOI}: Contact-Aware Humanoid-Object Interaction by Imitating Generated Videos without Task-Specific Training},
  author={Bi, Zhihai and Zhang, Qiang and Zhao, Guoyang and Cao, Jiahang and Luo, Xueyin and Zhang, Yushan and Xu, Jinglan and Geng, Ruoyu and Li, Yulin and Luo, Andrew F. and Ma, Jun},
  journal={arXiv:2606.12995},
  year={2026}
}

@article{xie2026grail,
  title={{GRAIL}: Generating Humanoid Loco-Manipulation from {3D} Assets and Video Priors},
  author={Xie, Tianyi and Zhang, Haotian and Park, Jinhyung and Wang, Zi and Wen, Bowen and Li, Jiefeng and Li, Xueting and Ben, Qingwei and Weng, Haoyang and Ye, Yufei and Minor, David and Wang, Tingwu and Jiang, Chenfanfu and Fidler, Sanja and Kautz, Jan and Fan, Linxi and Zhu, Yuke and Luo, Zhengyi and Iqbal, Umar and Yuan, Ye},
  journal={arXiv:2606.05160},
  year={2026}
}

@article{zhou2026exoactor,
  title={{ExoActor}: Exocentric Video Generation as Generalizable Interactive Humanoid Control},
  author={Zhou, Yanghao and Ma, Jingyu and Peng, Yibo and Sun, Zhenguo and Bai, Yu and Karlsson, B{\"o}rje F.},
  journal={arXiv:2604.27711},
  year={2026}
}

@article{chen2026imagine2real,
  title={{Imagine2Real}: Towards Zero-shot Humanoid-Object Interaction via Video Generative Priors},
  author={Chen, Jiahe and Wang, ZiRui and Jia, Feiyu and Chen, Xiao and Niu, Xiaojie and Zeng, Weishuai and Xue, Tianfan and Zhou, Xiaowei and Pang, Jiangmiao and Wang, Jingbo},
  journal={arXiv:2605.22272},
  year={2026}
}

@article{han2026video2sim2real,
  title={{Video2Sim2Real}: Full-Stack Autonomous Dexterous Skill Acquisition from a Single Human Video},
  author={Han, Yunhai and Qiu, Jianuo and Bai, Linhao and Xiao, Ziyu and Zeng, Zihang and Liu, Yangcen and Yang, Zhaodong and Jain, Shalin and Ma, Wenrui and Fu, Jiaqi and Zheng, Yuqian and Natarajan, Manisha and Irshad, Muhammad Zubair and Shaw, Kenneth and Gombolay, Matthew and Kira, Zsolt and Ravichandar, Harish},
  journal={arXiv:2606.08828},
  year={2026}
}

@article{he2026skip,
  title={{SKIP}: Sparse Keyframe Interpolation Paradigm for Efficient Embodied World Models},
  author={He, Ziheng and Chen, Yixiang and Yang, Ning and Wu, Zhanqian and Ma, Qisen and Xu, Yuan and Yang, Jiabing and Li, Peiyan and Wu, Xiangnan and Wang, Xiaofeng and Zhu, Zheng and Liu, Jing and Liu, Nianfeng and Huang, Yan},
  journal={arXiv:2606.00664},
  year={2026}
}

@article{ziakas2026grounding,
  title={Grounding Generated Videos in Feasible Plans via World Models},
  author={Ziakas, Christos and Bar, Amir and Russo, Alessandra},
  journal={arXiv:2602.01960},
  year={2026}
}

@article{kim2026robotfactored,
  title={Robot-Factored World Models via Robot Rendering},
  author={Kim, Byungjun and Kim, Taeksoo and Cha, Hyunsoo and Joo, Hanbyul},
  journal={arXiv:2607.22535},
  year={2026}
}

@inproceedings{chen2026craft,
  title={{CRAFT}: Video Diffusion for Bimanual Robot Data Generation},
  author={Chen, Jason and Liu, I-Chun Arthur and Sukhatme, Gaurav and Seita, Daniel},
  booktitle={IROS},
  year={2026}
}

@article{he2026vdreamer,
  title={{V-Dreamer}: Automating Robotic Simulation and Trajectory Synthesis via Video Generation Priors},
  author={He, Songjia and Chen, Zixuan and Ding, Hongyu and Shao, Dian and Shi, Jieqi and Li, Chenxu and Huo, Jing and Gao, Yang},
  journal={arXiv:2603.18811},
  year={2026}
}

@article{guo2026roboedit,
  title={{RoboEdit}: Turning Human Manipulation Videos into Scalable Robot Experience},
  author={Guo, Yaowei and Tao, Zeng and Jiang, Yuxin and Chen, Yunuo and Dou, Zhiyang and Ma, Yuxiang and Yang, Yin and Terzopoulos, Demetri and Jiang, Ying and Jiang, Chenfanfu},
  journal={arXiv:2608.18948},
  year={2026}
}

@inproceedings{karunratanakul2024dno,
  title={Optimizing Diffusion Noise Can Serve As Universal Motion Priors},
  author={Karunratanakul, Korrawe and Preechakul, Konpat and Aksan, Emre and Beeler, Thabo and Suwajanakorn, Supasorn and Tang, Siyu},
  booktitle={CVPR},
  pages={1334--1345},
  year={2024}
}

@article{rempe2026kimodo,
  title={{Kimodo}: Scaling Controllable Human Motion Generation},
  author={Rempe, Davis and Petrovich, Mathis and Yuan, Ye and Zhang, Haotian and Peng, Xue Bin and Jiang, Yifeng and Wang, Tingwu and Iqbal, Umar and Minor, David and de Ruyter, Michael and Li, Jiefeng and Tessler, Chen and Lim, Edy and Jeong, Eugene and Wu, Sam and Hassani, Ehsan and Huang, Michael and Yu, Jin-Bey and Chung, Chaeyeon and Song, Lina and Dionne, Olivier and Kautz, Jan and Yuen, Simon and Fidler, Sanja},
  journal={arXiv:2603.15546},
  year={2026}
}

@inproceedings{chen2025videodepthanything,
  title={{Video Depth Anything}: Consistent Depth Estimation for Super-Long Videos},
  author={Chen, Sili and Guo, Hengkai and Zhu, Shengnan and Zhang, Feihu and Huang, Zilong and Feng, Jiashi and Kang, Bingyi},
  booktitle={CVPR},
  year={2025}
}

@inproceedings{potamias2024wilor,
  title={{WiLoR}: End-to-End {3D} Hand Localization and Reconstruction In-the-Wild},
  author={Potamias, Rolandos Alexandros and Zhang, Jinglei and Deng, Jiankang and Zafeiriou, Stefanos},
  booktitle={CVPR},
  pages={12242--12254},
  year={2025}
}

@inproceedings{wen2024foundationpose,
  title={{FoundationPose}: Unified {6D} Pose Estimation and Tracking of Novel Objects},
  author={Wen, Bowen and Yang, Wei and Kautz, Jan and Birchfield, Stan},
  booktitle={CVPR},
  pages={17868--17879},
  year={2024}
}

@inproceedings{labbe2021robopose,
  title={Single-View Robot Pose and Joint Angle Estimation via Render \& Compare},
  author={Labb{\'e}, Yann and Carpentier, Justin and Aubry, Mathieu and Sivic, Josef},
  booktitle={CVPR},
  pages={1654--1663},
  year={2021}
}

@article{seedance2026,
  title={{Seedance 2.0}: Advancing Video Generation for World Complexity},
  author={{Seedance 2.0 Team}},
  journal={arXiv:2604.14148},
  year={2026}
}

@inproceedings{carion2025sam3,
  title={{SAM 3}: Segment Anything with Concepts},
  author={Carion, Nicolas and Gustafson, Laura and Hu, Yuan-Ting and Debnath, Shoubhik and Hu, Ronghang and Suris Coll-Vinent, Didac and Ryali, Chaitanya and Alwala, Kalyan Vasudev and Khedr, Haitham and Huang, Andrew and others},
  booktitle={ICLR},
  year={2026}
}

@article{jung2026tapnextpp,
  title={{TAPNext++}: What's Next for Tracking Any Point ({TAP})?},
  author={Jung, Sebastian and Zholus, Artem and Sundermeyer, Martin and Doersch, Carl and Goroshin, Ross and Tan, David Joseph and Chandar, Sarath and Triebel, Rudolph and Tombari, Federico},
  journal={arXiv:2604.10582},
  year={2026}
}

@misc{lumalabs2026ray32,
  author={{Luma AI}},
  title={{Ray 3.2}: Introduction and Core Concepts},
  year={2026},
  howpublished={\url{https://lumalabs.ai/learning-center/articles/ray-3-2-introduction-and-core-concepts}},
  note={Accessed September 1, 2026}
}

@article{xu2026dream2act,
  title={Morphology-Consistent Humanoid Interaction through Robot-Centric Video Synthesis},
  author={Xu, Weisheng and Li, Jian and Gu, Yi and Yang, Bin and Chen, Haodong and Lin, Shuyi and Zhou, Mingqian and Tan, Jing and Wu, Qiwei and Jiang, Xiangrui and others},
  journal={arXiv:2603.19709},
  year={2026}
}

@article{peng2018deepmimic,
  title={{DeepMimic}: Example-guided deep reinforcement learning of physics-based character skills},
  author={Peng, Xue Bin and Abbeel, Pieter and Levine, Sergey and Van de Panne, Michiel},
  journal={ACM Transactions On Graphics (TOG)},
  volume={37},
  number={4},
  pages={1--14},
  year={2018},
  publisher={ACM New York, NY, USA}
}

@article{luo2025sonic,
  title={{SONIC}: Supersizing motion tracking for natural humanoid whole-body control},
  author={Luo, Zhengyi and Yuan, Ye and Wang, Tingwu and Li, Chenran and Chen, Sirui and Castaneda, Fernando and Cao, Zi-Ang and Li, Jiefeng and Minor, David and Ben, Qingwei and others},
  journal={Sci. Robot.},
  volume={11},
  number={117},
  pages={eaed4592},
  year={2026}
}

@article{yin2025unitracker,
  title={{UniTracker}: Learning universal whole-body motion tracker for humanoid robots},
  author={Yin, Kangning and Zeng, Weishuai and Fan, Ke and Dai, Minyue and Wang, Zirui and Zhang, Qiang and Tian, Zheng and Wang, Jingbo and Pang, Jiangmiao and Zhang, Weinan},
  journal={arXiv:2507.07356},
  year={2025}
}

@article{liao2025beyondmimic,
  title={{BeyondMimic}: From motion tracking to versatile humanoid control via guided diffusion},
  author={Liao, Qiayuan and Truong, Takara E and Huang, Xiaoyu and Gao, Yuman and Tevet, Guy and Sreenath, Koushil and Liu, C Karen},
  journal={Sci. Robot.},
  volume={11},
  number={117},
  pages={eadx8924},
  year={2026}
}

@inproceedings{he2024omnih2o,
  title={{OmniH2O}: Universal and dexterous human-to-humanoid whole-body teleoperation and learning},
  author={He, Tairan and Luo, Zhengyi and He, Xialin and Xiao, Wenli and Zhang, Chong and Zhang, Weinan and Kitani, Kris and Liu, Changliu and Shi, Guanya},
  booktitle={CoRL},
  year={2024}
}

@inproceedings{fu2025humanplus,
  title={{HumanPlus}: Humanoid Shadowing and Imitation from Humans},
  author={Fu, Zipeng and Zhao, Qingqing and Wu, Qi and Wetzstein, Gordon and Finn, Chelsea},
  booktitle={CoRL},
  pages={2828--2844},
  year={2025},
  organization={PMLR}
}

@inproceedings{ze2025twist2,
  title={{TWIST2}: Scalable, portable, and holistic humanoid data collection system},
  author={Ze, Yanjie and Zhao, Siheng and Wang, Weizhuo and Kanazawa, Angjoo and Duan, Rocky and Abbeel, Pieter and Shi, Guanya and Wu, Jiajun and Liu, C Karen},
  booktitle={ICRA},
  year={2026}
}

@inproceedings{zhang2024wococo,
  title={{WoCoCo}: Learning whole-body humanoid control with sequential contacts},
  author={Zhang, Chong and Xiao, Wenli and He, Tairan and Shi, Guanya},
  booktitle={CoRL},
  year={2024}
}

@article{weng2025hdmi,
  title={{HDMI}: Learning interactive humanoid whole-body control from human videos},
  author={Weng, Haoyang and Li, Yitang and Sobanbabu, Nikhil and Wang, Zihan and Luo, Zhengyi and He, Tairan and Ramanan, Deva and Shi, Guanya},
  journal={arXiv:2509.16757},
  year={2025}
}

@inproceedings{yang2025omniretarget,
  title={{OmniRetarget}: Interaction-preserving data generation for humanoid whole-body loco-manipulation and scene interaction},
  author={Yang, Lujie and Huang, Xiaoyu and Wu, Zhen and Kanazawa, Angjoo and Abbeel, Pieter and Sferrazza, Carmelo and Liu, C Karen and Duan, Rocky and Shi, Guanya},
  booktitle={ICRA},
  year={2026}
}

\end{document}